\pdfoutput=1

\documentclass[11pt]{article}

\usepackage[]{ACL2023}
\usepackage{todonotes}

\usepackage{fancyvrb}

\usepackage{listings}

\usepackage{times}
\usepackage{latexsym}

\usepackage{amsmath}
\usepackage{graphicx}
\usepackage{multirow}
\usepackage{booktabs}
\usepackage{fancyvrb}
\usepackage{setspace}
\usepackage{xcolor}

\usepackage[T1]{fontenc}

\usepackage[utf8]{inputenc}

\usepackage{microtype}

\usepackage{inconsolata}

\title{ChatGPT for Zero-shot Dialogue State Tracking:\\A Solution or an Opportunity?}

\author{Michael Heck, Nurul Lubis, Benjamin Ruppik, Renato Vukovic,\\\textbf{Shutong Feng, Christian Geishauser, Hsien-Chin Lin, Carel van Niekerk, Milica Ga\v{s}i\'{c}} \\
  Heinrich Heine University Düsseldorf, Germany \\
  \texttt{\{heckmi,lubis,ruppik,revuk100,fengs,geishaus,linh,niekerk,gasic\}@hhu.de}}

\begin{document}
\maketitle
\setstretch{0.99} 

\begin{abstract}
Recent research on dialogue state tracking (DST) focuses on methods that allow few- and zero-shot transfer to new domains or schemas. However, performance gains heavily depend on aggressive data augmentation and fine-tuning of ever larger language model based architectures.
In contrast, general purpose language models, trained on large amounts of diverse data, hold the promise of solving any kind of task without task-specific training. We present preliminary experimental results on the ChatGPT research preview, showing that ChatGPT achieves state-of-the-art performance in zero-shot DST. Despite our findings, we argue that properties inherent to general purpose models limit their ability to replace specialized systems.
We further theorize that the in-context learning capabilities of such models will likely become powerful tools to support the development of dedicated and dynamic dialogue state trackers.
\end{abstract}

\section{Introduction}

Dialogue state tracking (DST) is a critical component for task-oriented dialogue systems. Its purpose is to extract and track user's goals throughout a conversation~\cite{young2010hidden}. DST is challenging due to the infinite possibilities of user/agent conversations, and because services and schemas/APIs that dialogue systems interface are subject to constant change~\cite{ren2018towards}. Although traditional approaches achieve high accuracy when operating on a pre-defined set of concepts called an ontology~\cite{mrkvsic2016neural,liu2017end,zhong2018global}, 
ongoing research explores transfer to new domains with little to no additional learning~\cite{rastogi2020schema} using ontology independent architectures to allow seamless adaptation to out-of-ontology concepts.

Many strategies for zero-shot transfer to unseen domains have been proposed.
\citet{li2021zero} treat DST as a question answering (QA) task by leveraging data augmentation.
\citet{zhao2022description} propose DST by relying on schema descriptions while~\citet{heck2022robust} utilize natural language descriptions to facilitate zero-shot transfer.
\citet{gao-etal-2020-machine} and \citet{lin2021zero} suggest learning from non-dialogue QA data which are available in large amounts to improve generalization.
\citet{campagna2020zero} harness large synthesized data based on abstract dialogue models.
However, none of these techniques are ideal solutions. Fine-tuning
is challenging due to computational costs, risk of over-fitting and the need for expensive~\cite{budzianowski2018multiwoz} task-specific data.
Cross-task transfer still requires curated data
and careful consideration of suitable learning tasks.
Data augmentation requires high level task knowledge and an adequate synthesizing strategy. 

A new generation of large language models (LLMs)~\cite{brown2020gpt3paper,ouyang2022training,glaese2022improving} comes with the promise to be equipped to solve any task without task-specific fine-tuning, but solely
with world knowledge they acquired during self-training on massive amounts of data.
Such LLMs have been shown to perform remarkably well on in-context learning (ICL), where only a natural language prompt and examples are provided to condition the generation process, achieving significant improvements over fine-tuned approaches in few-shot setups~\cite{brown2020gpt3paper,wang2022benchmarking}. 
ChatGPT~\cite{chatgpt_blog_openai_2023} -- trained using human feedback and reinforcement learning -- is the most recent of such models and single-handedly solves an array of challenging natural language processing (NLP) tasks with super-human capabilities, 
all through a natural language dialogue interface.

In this work, we aim to answer the question: \emph{does ChatGPT solve the problem of zero-shot DST?}
We show 
that crafting intuitive natural language prompts is sufficient to achieve state-of-the-art performance with ChatGPT, exceeding conventional, engineering-heavy approaches to zero-shot DST by a large margin.
However, despite our findings, we argue
that properties inherent to general purpose models inhibit their ability to simply replace specialized systems. 
We speculate that while in the foreseeable future general purpose models may not become holistic solutions to complex problems, they will provide ample opportunities to empower specialized systems to go beyond their pre-defined scopes, enable on-the-fly extensibility and generation of high quality training data by zero-shot synthesizing or automatic labeling.

\section{Background}
\label{sec:background}

\paragraph{Dialogue state tracking} is tasked to (1) determine for every turn $t$ in a dialogue $\{(U_t, M_t)\}_1^T$ with $U_t$ and $M_t$ being current user and preceding system utterance whether any of the slots in $S = \{S_n\}_1^N$ is present, to (2) predict values for each $S_n$ and to (3) track the dialogue state $\mathit{DS}_t$ $\forall t \in [1, T]$.
The DS is cumulative, i.e., $\mathit{DS}_t = \mathrm{update}(\mathit{DS}_{t-1}, \widehat{\mathit{DS}}_{t})$ is updated given the predictions of slot-value updates $\widehat{\mathit{DS}}_{t}$.

\paragraph{ChatGPT}~\cite{chatgpt_blog_openai_2023} is a dialogue agent~\cite{Leike2018ScalableAA}, and in its core a GPT-3.5 LLM fine-tuned on human-written prompt-response pairs followed by reinforcement learning with human feedback (RLHF)~\cite{christiano2017RLHF, stiennon2020humanfeedback}. 
RLHF utilizes a reward model trained on human feedback to improve generation quality and adequacy via proximal policy optimization~\cite{Schulman2017ProximalPO}, thereby \emph{aligning} model output to human values and user's expectations.
At the time of writing this work, ChatGPT is proprietary.
As a sibling model to InstructGPT, details of its training are elaborated by \citet{ouyang2022training}.

\section{Zero-shot DST with ChatGPT}

Our investigative approach to zero-shot DST with ChatGPT differs considerably from related works. We decode dialogue state updates with a general purpose model, without undergoing any parameter updates. Consequently, we neither employ data augmentation nor cross-task transfer learning. Instead, we solely rely on the general capacities of ChatGPT as an aligned dialogue agent.
We take a most rigorous approach to zero-shot transfer where we do not allow the provision of any examples, nor of a formal task definition. 
Instead, we only permit natural language explanations of what the model is supposed to do. This sets our investigation apart from the closely related IC-DST~\cite{hu2022incontext}.

In zero-shot DST, the set of slots $S$ relevant during inference and the set of slots $S^{\prime}$ seen during training of the model $X_{\theta}$ with parameters $\theta$ are disjoint, i.e., $S \cap S^{\prime} = \emptyset$.
Further, it may be $S^{\prime} = \emptyset$, in which case $\theta$ is not specifically tuned towards solving DST. This is precisely the case for ChatGPT in our setup.
Our approach to zero-shot DST with ChatGPT is formalized as follows. Let
\begin{equation*}
  \begin{split}
    A_1 =& P \oplus \mathrm{"system"}{:} M_1 \oplus \mathrm{"user"}{:} U_1, \\
    A_t =&  \mathrm{"system"}{:} M_t \oplus \mathrm{"user"}{:} U_t, \forall t \in [2,T],    
  \end{split}
\end{equation*}
where $P$ is the task description which provides the model with instructions for how to process a dialogue between a system $M$ and a user $U$. $A_1$ is the initial prompt to ChatGPT. $A_{t \ge 2}$ are the follow-up prompts, only containing a single turn-pair of the dialogue of interest. ChatGPT is particularly suitable for this strategy due to its chat based interface.
ChatGPT generates its next output $B_{t}$ conditioned on the current prompt $A_{t-1}$ , as well as all preceding user queries and system responses of the same chat. 
The dialogue state update $\widehat{DS}_{t}$ can be found in $B_{t}$, but may not be directly interpretable as such due to the diversity in the output surface forms. Thus, we require a normalization operation $\widehat{DS}_{t} = \mathrm{normalize}(B_{t})$.
In contrast to~\cite{hu2022incontext}, we do not condition $B_{t}$ on ${DS}_{t}$. This renders the task even more challenging, as ChatGPT is forced to solve complex subtasks such as coreference resolution -- the case where a newly encountered slot refers to the value of another slot -- solely given the initial prompt and its own latent dialogue state given the dialogue history.

\section{Experiments}

At the time of conducting our experiments, ChatGPT is a proprietary research preview accessible for free via a web interface\footnote{\url{chat.openai.com} (accessed 6. Jan. to 20. Jan. 2023)}. We used the \texttt{Jan 9} version
of the model. 
We use a regular expression term to extract all parts that are JSON formatted.
We form $DS_t$ by accumulating all predicted updates up to turn $t$.

\paragraph{Evaluation.}
We evaluate on the 1000 dialogues of the MultiWOZ 2.1~\cite{eric2019multiwoz} test split and use joint goal accuracy (JGA) to compare methods. 
For a fair judgement of the ChatGPT predictions, we follow the evaluation procedure of~\citet{heck-etal-2020-trippy}. 
We process each dialogue once and refrain from using ChatGPT's \emph{regeneration} feature.

\paragraph{Prompt.}
We imposed restrictions that the task-defining prompt $P$ be intuitive natural language and provides no formal schema. The crafting process involves simple trial-and-error on fewer than 10 held-out dialogues from the MultiWOZ training set. 
The design process was guided by the intention to imitate the behavior of a triple copy strategy (TripPy) DST~\cite{heck-etal-2020-trippy}. $P$ consists of three parts. First, a list of names for detectable informable slots along with natural language descriptions. 
The slot names help us extract a $\widehat{DS}_t$ that is compatible with the dataset's labels. Second, a sparse list of slots that are categorical, along with their value candidates
for (1) aiding normalization of values that are expected to show high variability in expression, and (2) modeling Boolean slots. 
Third, an informal task description.\footnote{See Appendix~\ref{app:prompt} for the full prompt.}

\begin{figure}[t]
	\centering
	\includegraphics[page=3, trim=6.5cm 5.3cm 6.3cm 5.6cm, clip=true, width=1.00\linewidth,]{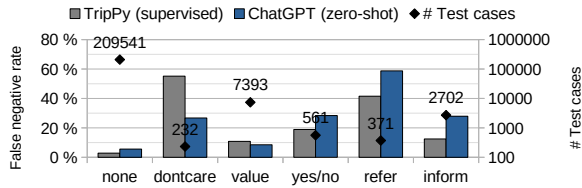}
	\caption{Missed predictions by type of value.}
	\label{fig:classtypes}
\end{figure}

\subsection{ChatGPT vs. Supervised SOTA}
\label{sec:experiments:ssec:performance}

Comparing ChatGPT's performance to state-of-the-art \emph{supervised} approaches that achieve close to 60\% JGA is not a fair fight\footnote{\url{https://github.com/budzianowski/multiwoz}}, and yet we observe an impressive 31.5\% zero-shot JGA. This result is double-edged; on the one hand it is evidence that ChatGPT
is capable of DST\footnote{See Appendix~\ref{app:example} for an example dialogue.}, and on the other hand is no match for specialized systems.

The comparison to TripPy, a SOTA supervised model, allows us a more fine-grained analysis. In Figure~\ref{fig:classtypes}, slot filling performance is broken down into value types. We observed that ChatGPT underperforms in non-trivial cases, namely \emph{refer}, where a newly encountered slot refers to the value of another slot, and \emph{inform}, where a slot-value was mentioned by the system and confirmed by the user. ChatGPT shows slight underperformance for Boolean slots. 
Remarkably, performance for values that are extracted directly from user utterances -- the most relevant category in terms of frequency -- is exceeding the strong supervised baseline. Lastly, ChatGPT has a clear advantage in the underrepresented and therefore notoriously difficult \emph{dontcare} cases, where a user is indifferent about a particular value for a slot.

\begin{table}[t]
  \setlength{\tabcolsep}{2pt}
  \centering
  \footnotesize
  \begin{tabular}{@{}l|ccccc|c@{}}
    \toprule
    Models & attr. & hotel & rest. & taxi & train & avg. \\
    \midrule
    TRADE~\citeyearpar{wu2019transferable,campagna2020zero} & 22.8  & 19.5  & 16.4  & 59.2  & 22.9  & 28.16 \\
    TripPy-R~\citeyearpar{heck2022robust} & 27.1  & 18.3  & 15.3  & 61.5  & 23.7  & 29.18 \\
    TransferQA~\citeyearpar{lin2021zero} & 31.3  & 22.7  & 26.3  & 61.9  & 36.7  & 35.78 \\
    \citet{li2021zero} & 42.4  & 24.9  & 27.7  & 60.3  & 41.1  & 39.28 \\
    D3ST~\citeyearpar{zhao2022description} & \textbf{56.4}  & 21.8  & 38.2  & \textbf{78.4}  & 38.7  & 46.70 \\
    \citet{campagna2020zero} & 52.8  & 36.3  & 45.3  & 62.6  & 46.7  & 48.74 \\
    \midrule
    ChatGPT & 52.7  & \textbf{42.0}  & \textbf{55.8}  & 70.9  & \textbf{60.8}  & \textbf{56.44} \\
    \midrule\midrule
    IC-DST\footnotemark[5]~\citeyearpar{hu2022incontext} & 60.0  & 46.7  & 57.3  & 71.4  & 49.4  & 56.96 \\ 
    \bottomrule
  \end{tabular}
  \caption{Performance comparison between state-of-the-art methods for zero-shot DST in per-domain JGA.}
  \label{tab:zeroshot}
\end{table}

\subsection{ChatGPT vs. Zero-shot SOTA}
\label{sec:experiments:ssec:zeroshot}

ChatGPT considerably outperforms previous approaches to zero-shot DST (see Table~\ref{tab:zeroshot}) and is more stable across domains than other methods. The model tends to handle challenging domains markedly better, while maintaining high performance on domains that are handled with relative ease by earlier approaches.
Most approaches to zero-shot DST still employ supervised learning on a subset of domains and test on a held-out domain. Such methods struggle in domains with many slots never seen during training. This is evident for \emph{hotel}, which has many unique slots and is the only domain with Boolean slots. ChatGPT can excel in such challenging scenarios by drawing from its general world knowledge to interpret concepts. \emph{taxi} is challenging due to its frequent \emph{refer} cases. Where most other methods fail, ChatGPT shows competency in resolving co-references in the zero-shot setting.
Other models designed for DST rely on architectures that are not fundamentally different from the backbone model of ChatGPT. The reason for ChatGPT's superior abilities in conducting DST is likely found in its training scheme, particularly instruction tuning and alignment via reinforcement learning with human feedback~\cite{ouyang2022training,DBLP:journals/corr/abs-1909-08593}, combined with its massive scale in terms of model and training data size.
IC-DST~\cite{hu2022incontext} was the first successful attempt at pseudo\footnote{\label{fn:1}\citet{hu2022incontext} uses hand-crafted labeled examples for ICL even in the ``zero-shot'' case.} zero-shot DST via ICL. Our preliminary results with ChatGPT are on par, which is remarkable for the following reasons. (1) Our prompt is non-schematic and without examples,
(2) our task-defining prompt
is stated only once at the beginning of the chat,
and (3) we do not maintain a DS to serve as additional input at each turn. 
The heightened zero-shot performance of IC-DST can be mainly attributed to these points.

\begin{table*}[t]
  \setlength{\tabcolsep}{2pt}
  \centering
  \footnotesize
  \begin{tabular}{@{}lll@{}}
    \toprule 
    a) & PMUL4050 & system: "I'd recommend the Autumn House. Would you like to make a booking?" \\
    & & user: "Yes please. I need the reservation to be for 8 people and 2 nights starting on Tuesday." \\
    & & \textbf{Prediction:} {\color{blue}... \color{red}hotel-name: none}
      \textbf{Label:} {\color{blue}..., hotel-name: autumn house} \\
    \midrule 
    b) & PMUL0117 & user: "Yes I also need a taxi that will get me to the restaurant by the booked time please." \\
    & & \textbf{Prediction:} {\color{red}taxi-destination: hotel, taxi-departure: restaurant} \\
    & & \textbf{Label:} {\color{blue}taxi-destination: the gonville hotel, taxi-departure: la mimosa} \\
    \midrule 
    c) & SNG01873 & user: "I need to be picked up from pizza hut city centre after 04:30" \\
    & & \textbf{Prediction:} {\color{blue}..., \color{red}hotel-name: dontcare, ..., attraction-type: dontcare, ...} 
      \textbf{Label:} {\color{blue}...} \\
    \midrule 
    d) & PMUL0599 & user: "[...] Can you just help me find a high-end Mexican restaurant?" \\
    & & \textbf{Prediction:} {\color{blue}..., \color{red}restaurant-pricerange: high-end}
      \textbf{Label:} {\color{blue}..., restaurant-pricerange: expensive} \\
    \midrule 
    e) & MUL2051 & user: "Can I get address and postcode for the hotel?" \\
    & & \textbf{Prediction:} {\color{red}hotel-address: ?, hotel-postcode: ?}
      \textbf{Label:} {\color{blue}-} \\
    & & system: "The address is 74 chesterton road, the postal code is cb41er, can I assist with anything else?" \\
    & & user: "That is all for now, goodbye." \\
    & & \textbf{Prediction:} {\color{red}hotel-address: 74 chesterton road, hotel-postcode: cb41er}
      \textbf{Label:} {\color{blue}-} \\
    \midrule 
    f) & MUL0524 & user: "I'm going to Cambridge on saturday and want to arrive by 14:15 please." \\
    & & \textbf{Prediction:} {\color{blue}..., \color{red}train-day: Saturday}
      \textbf{Label:} {\color{blue}..., train-day: saturday} \\
    \midrule 
    g) & PMUL4246 & user: "i need a place to go and should be a museum" \\
    & & \textbf{Prediction:} {\color{blue}attraction-type: museum}
      \textbf{Label:} {\color{blue}attraction-type: museum} \\
    & & system: "Okay! There are several museums in Cambridge. What part of town would you like to visit?" \\
    & & user: "How about ones in the centre, what's available?" \\
    & & \textbf{Prediction:} {\color{red}attraction-type: museum, \color{blue}attraction-area: centre}
      \textbf{Label:} {\color{blue}attraction-area: centre} \\
    \bottomrule
  \end{tabular}
  \caption{Examples of reoccurring errors as listed in Section~\ref{sec:experiments:ssec:errors}.}
  \label{tab:examples}
  \vspace{-3pt}
\end{table*}

\subsection{Error Analysis}
\label{sec:experiments:ssec:errors}

We identified a set of recurring errors that are likely caused by either the content of $P$ or by the model's inherent properties.
See Table~\ref{tab:examples} for examples. 
See also Appendix~\ref{app:errors} lists for more detailed instances.

\paragraph{a) Failed carry-over of system-informed values.} 
Our $P$ does not explicitly instruct to resolve \emph{inform} cases (see Section~\ref{sec:experiments:ssec:performance}) cases. Nevertheless, ChatGPT handles the majority of cases correctly, failing to carry over only about 28\% of system-informed values. Specifying the desired behavior in $P$ may improve this ratio further.

\paragraph{b) Incomplete coreference resolution.} Coreferences are usually detected -- i.e., in about 65\% of cases --, but often not resolved. Where a coference was detected, about 23\% are not resolved correctly, and another 13\% are incorrect due to other errors.

\paragraph{c) Overprediction of \emph{dontcare}.} The recall of ChatGPT for \emph{dontcare} is considerably higher than for the supervised baseline, but precision is low.
About 35\% of \emph{none} false negatives in Figure~\ref{fig:classtypes} can be attributed to overpredicting \emph{dontcare} by ChatGPT, compared to 5\% for the supervised baseline.
This is likely caused by the formulation in $P$. Occasionally, the model interprets slots that are not specifically filled by the user as \emph{dontcare}.

\paragraph{d) Ignoring value candidates.} On rare occasions, ChatGPT ignores value candidates for categorical slots and picks variants from the dialogue context instead. We observed this error for 0.1\% of all values to be extracted from the context.

\paragraph{e) Hallucinated slots.} The model frequently hallucinates slots. 
About 90.4\% of all of ChatGPT's slot predictions are MultiWOZ slots. Since we specifically prompt ChatGPT to fill slots requested by the user with ``?'', the vast majority of hallucinations -- 8.6\% of all slot predictions -- are of the \emph{requestable} type, which are not considered by the standard MultiWOZ evaluation and are therefore not listed in $P$. In fact, ChatGPT predicts all requestable slots appearing in the MultiWOZ dataset with an average recall of 61\%. Rarely -- in 0.3\% of all cases -- alternative names are hallucinated for slots listed in $P$. A further 0.6\% are predictions for made up slots.

\paragraph{f) Arbitrary normalization.} We observed that the model sometimes chooses to normalize predicted values. However, these normalizations are inconsistent across dialogues.

\paragraph{g) Predicting $DS_t$ instead of $\widehat{DS}_t$.} Despite explicitly requesting to predict DS updates, ChatGPT on rare occasions -- in 0.2\% of all processed dialogues -- attempts to predict the full DS at each turn, which may cause other phenomena such as slot-value over-prediction.

\section{Discussion: Solution or opportunity?\footnote{The model's own response on that matter is refreshingly balanced. See Appendix~\ref{app:question} for ChatGPT's response.}}

ChatGPT is a sophisticated dialogue agent that, via alignment with human judgements, is capable of understanding context and intent of a multi-turn conversation far beyond the capacities of the previous generation of LLMs. This makes it well-suited for DST. 
Our results demonstrate that even with intuitive natural language prompts, a complex task such as DST can be solved exceedingly well without any form of additional learning. 

While specialized systems can exert control over its input-processing and output-generation to arbitrary degrees, this is not the case for ChatGPT. Even with the most rigorous and schematic prompts, there can be no guarantee that the model interprets the input as intended or generates the output as required, which may lead to unexpected behavior. Furthermore, there is no guarantee that behavior is consistent across a series of similar inferences, such as in our experimental evaluation.
In terms of deployment, the cost factor of building and running massive models may hinder their utilization as a plug-and-play module. 

Despite impressive zero-shot and ICL results for general purpose models, specialist models still perform best on most tasks thanks to task-specific solutions via adequate engineering~\cite{heck-etal-2020-trippy, ye2021slot, kim2019efficient} and task-related data. However, the opportunities to improve dedicated systems with the help of general purpose models are plenty. Their predictive powers could be used for developing smaller, specialized, low inference cost models. Automatic labeling and data augmentation are natural use cases for ChatGPT, as is evident from our experimental results; a per-domain JGA of 70\% (see Section~\ref{sec:experiments:ssec:zeroshot}) is surely sufficient to generate additional mid- to high-quality training data for dedicated systems. Automatic labeling may be conducted on-line for on-the-fly adaptation of production systems or off-line for iterative learning.

Another way of harnessing general purpose models is the integration into dedicated systems as fallback options in case of out-of-domain or out-of-ontology requests. An integration via knowledge-seeking term detection~\cite{DBLP:journals/corr/abs-2011-06486} could facilitate the ability to provide context-aware responses that go beyond the original scope of the specialized system. General purpose models may handle unseen domains in place of the main model.

While hallucinations may be an issue if not handled adequately, they also pose an opportunity to enable zero-shot concept detection. We observed that many slot hallucinations were sensible and pointed at elements that were meaningful to conversations. Zero-shot slot detection may be utilized to annotate and prepare unstructured data for model training, and to expand a system's capacities on-the-fly. Dialogue state trackers with dynamic dialogue states have the potential to expand a task-oriented dialogue system's conversational range seamlessly~\cite{geishauser-etal-2022-dynamic}. A general purpose model that has the capacity to identify new concepts may be utilized to generate API calls and database queries that are unknown to the specialized system~\cite{chatgpt_plugins_openai_2023,langchain_2023}.

General purpose models may replace some components in a modular dialogue system~\cite{DBLP:journals/corr/abs-2211-17148}. It might still be beneficial to rely on specialized DST and a dedicated policy for particular tasks in order to maintain interpretability and a desired level of control over information flow. However, natural language understanding (NLU) and natural language generation (NLG) modules may be powered by generative large language model based systems such as ChatGPT in order to benefit from a heightened ability of semantic modeling and to facilitate more natural and diverse output, thus promoting more natural conversations with modular task-oriented dialogue systems.

\section{Conclusion}

This work is the first to investigate ChatGPT's capacities for zero-shot DST. Despite remarkable preliminary results that we achieved, we identified limitations rooted in inherent properties of general purpose models, preventing these to become holistic solutions to complex NLP problems without further research. We discussed opportunities provided by ChatGPT and similar models to advance the development of specialized systems. With our insights and discussion, we hope to stimulate research in similar directions.

\section*{Limitations}

At the time of writing this work, ChatGPT is only available as a proprietary free research preview via a web interface. This is limiting in several ways. (1) Parts of our analysis are qualitative, as quantification is challenging due to limited accessability of the investigated model. (2) Some details about the investigated model are not yet disclosed. This is true for the model design as well as for the data used to train ChatGPT. MultiWOZ is a freely available and widely used dataset, therefore no guarantee can be given that ChatGPT has not been exposed to at least some meta details regarding this dataset. (3) Given the nature of the free research preview, exact reproducibility is not guaranteed, as the model may change any time. However, it is expected that any future version of ChatGPT retains its general abilities and behaviors.

\paragraph{Model-as-a-service.}
Building a general purpose model such as ChatGPT is extremely costly and an option only for few. However, once it exists, it may be utilized for a multitude of purposes. As a model, ChatGPT does not need to be built for DST in order to be useful for DST. With capable enough general purpose models, fine-tuning towards specific tasks may be avoided. Fine-tuning is challenging for multiple reasons such as the need for adequate data, computational costs, risk of over-fitting and catastrophic forgetting, among others.

Just like its sibling model, ChatGPT will become available as model-as-a-service. The advantage of this is that a massive LM such as this is usable independent of the user's hardware. But this advantage comes with the disadvantage that it will in all probability remain proprietary. In consequence, it will likely not be possible to ever run, adapt, train or modify ChatGPT on local machines.

ChatGPT as model-as-a-service is likely to remain a black box to customers and researchers, even if just in parts. The model may change any time. In fact, a model update during our experimental evaluation prompted us to re-process a few of our test dialogues. This property impedes backward compatibility and the ability to trust in familiar behavior.

A general purpose model may show too general behavior and converse about more than what is required or requested. This also poses vulnerabilities for adversarial attacks. To this end, models such as ChatGPT have been trained with human feedback to better handle malicious intent and abusive behaviors.

A model-as-a-service is a gated resource. As such, its indefinite availability cannot be guaranteed. Further, recurring costs for access may be too high for certain downstream tasks. As a hosted service, latency might become a bottleneck or hindrance for its use as a component in complex applications.

\section*{Ethics Statement}

The disclaimer of ChatGPT states that the model may occasionally generate incorrect information and may occasionally produce harmful instructions or biased content.
Models, code and datasets were used in accordance with their respective licenses, terms of use and intended use.
We provide logs and code that we created for this work.\footnote{\url{https://gitlab.cs.uni-duesseldorf.de/general/dsml/chatgpt-dst-public}}
Data that we used and generated does not contain any information that names or uniquely identifies individual people or offensive content.

\section*{Acknowledgements}

M. Heck, N. Lubis, S. Feng and C. van Niekerk are supported by funding provided by the Alexander von Humboldt Foundation in the framework of the Sofja Kovalevskaja Award endowed by the Federal Ministry of Education and Research, while C. Geishauser, H-C. Lin, B. Ruppik and R. Vukovic are supported by funds from the European Research Council (ERC) provided under the Horizon 2020 research and innovation programme (Grant agreement No. STG2018804636). 
We thank Girish Kulkarni and Annika Hennes for their help in processing MultiWOZ dialogues with ChatGPT.

\bibliography{anthology,custom}
\bibliographystyle{acl_natbib}

\appendix

\section{Task-defining Prompt $P$}
\label{app:prompt}

See Figure~\ref{fig:unabridged_prompt_P} for our complete prompt.

\section{Example Dialogue}
\label{app:example}

\begin{figure*}
    \centering
    \includegraphics[trim=0.0cm 0.0cm 0.0cm 0.0cm, clip=true, width=1.00\linewidth]{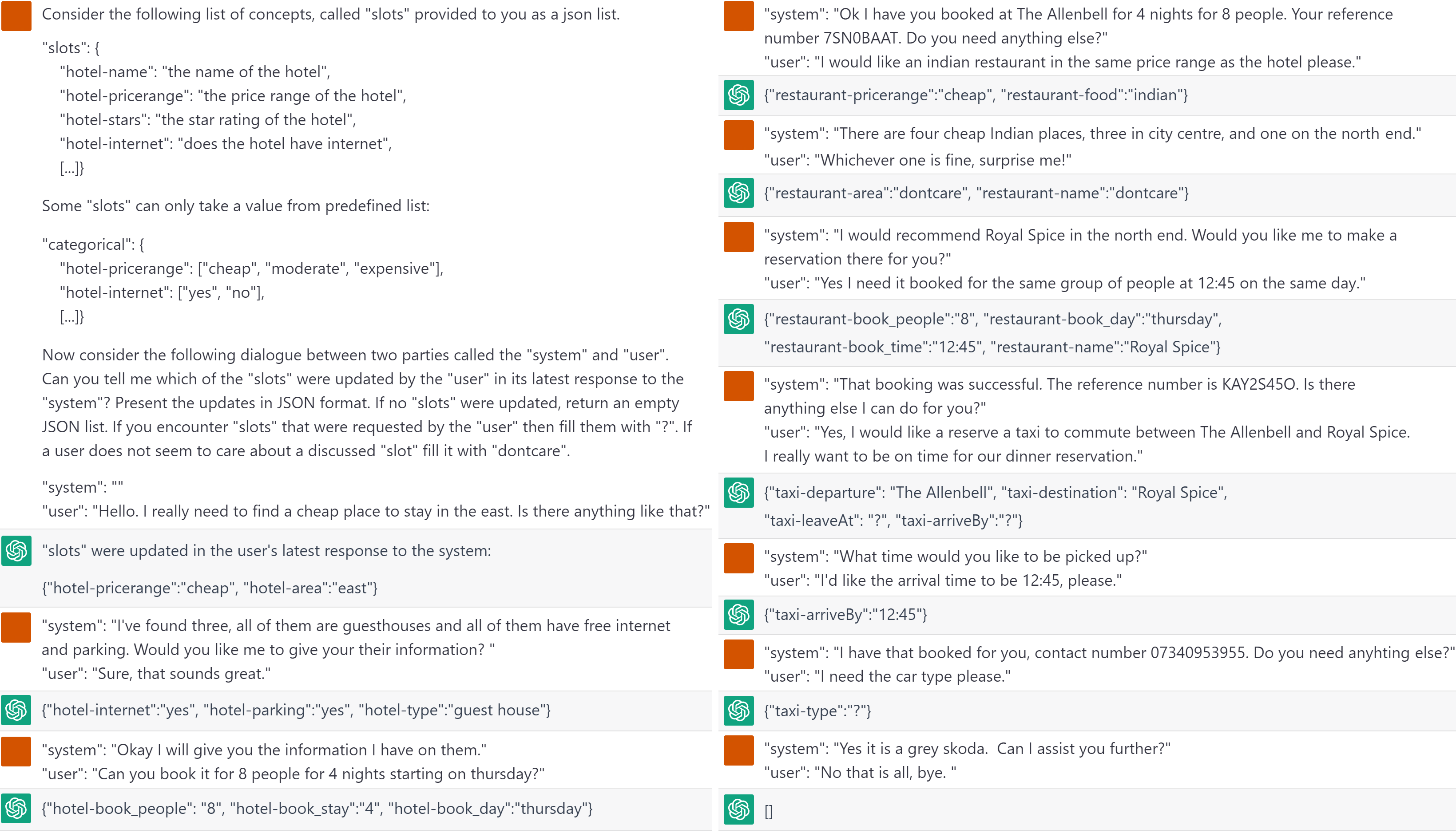}
    \caption{Example of DST with ChatGPT. $P$ is abridged for brevity. The sentence ``"slots" were updated [...]'' was generated at each turn. In turn 2, Boolean slots are predicted correctly, and a value variant (``guesthouse'') is mapped to the correct label. In turn 4 a coreference is resolved correctly. In turn 5, \emph{dontcare} is handled correctly. In turn 7, the model assumes implicit requests. In turn 9, ChatGPT hallucinates a slot which is not listed in $P$, but sensible.}
	\label{fig:example}
\end{figure*}

See Figure~\ref{fig:example} for an example dialogue that processes \texttt{MUL1350} for DST.

\subsection{Reasoning about Numbers}

From \texttt{MUL2122}:

\lstset{basicstyle=\ttfamily\tiny, breaklines=true, postbreak=\mbox{\textcolor{red}{$\hookrightarrow$}\space}}
\begin{lstlisting}
"system": ""
"user": "Hello!  I am looking for train schedules from Leicester.  Can you help me with that?"

The following slots were updated by the user:
{
"train-departure": "Leicester",
"train-destination": "?",
"train-day": "?",
"train-leaveAt": "?",
"train-arriveBy": "?",
"train-book_people": "?"
}

"system": "Sure! There are several trains leaving from leicester. What destination, day and time would you like me to search?"
"user": "I like a train going to cambridge on tuesday please."

The following slots were updated by the user:
{
"train-destination": "cambridge",
"train-day":"tuesday",
"train-leaveAt":"dontcare",
"train-arriveBy":"dontcare"
}

"system": "Okay.  It looks as if there are trains leaving Leicester for Cambridge every hour starting at 5:09 and ending at 23:09 on Tuesday. "
"user": "Pick one. I need the travel time and arrival time. Any one after 20:15."

The following slots were updated by the user:
{
"train-leaveAt":"dontcare",
"train-arriveBy":"20:15"
}

"system": "You have one leaving at 21:09, what time do you want to arrive by?"
"user": "Thank you that will work fine for me and my husband"

The following slots were updated by the user:
{"train-book_people":"2"}

"system": "The booking was successful. Your reference number is RWRKP9L6"
"user": "I also need a hotel with free wifi and free parking. "

The following slots were updated by the user:
{
"hotel-parking":"yes",
"hotel-internet":"yes"
}

"system": "I have several excellent options. Did you have an area of town in mind?"
"user": "No, I just would like a guesthouse. "

The following slots were updated by the user:
{
"hotel-type":"guest house",
"hotel-area":"dontcare"
}

"system": "Sure, I'd suggest Acorn Guest House at 154 Chesterton Road CB41DA. It's moderately priced with 4 stars. Would that work for you?"
"user": "Can you please make a reservation for just me, for 4 nights starting tuesday?"

The following slots were updated by the user:
{
"hotel-name":"Acorn Guest House",
"hotel-book_day":"Tuesday",
"hotel-book_stay":"4",
"hotel-book_people":"1"
}

"system": "Your reference number is HYEDH9O5, anything else I can help you with?"
"user": "No thank you, you have been very helpful. Have a great day."

An empty JSON list, no new slot is updated by the user.
\end{lstlisting}

\subsection{Explaining a Response Referring to $P$}

From \texttt{MUL2405}:
\begin{lstlisting}
"system": ""
"user": "Hi. I am having trouble finding a place to eat called prezzo. Can you locate it for me and tell me a little about it please?"
There were no "slots" updated in the "user"'s latest response to the "system". As per the provided user sentence, the user is seeking information and is not providing any details. Thus the response should be an empty JSON list {}

"system": "Prezzo is located at 21 - 24 Northampton Road, in the west. It's an italian restaurant. Moderate price range."
"user": "That sounds perfect!"
The "slots" updated by the "user" in this response are the following:

Copy code
{
    "restaurant-name": "prezzo",
    "restaurant-area": "west",
    "restaurant-food": "italian",
    "restaurant-pricerange": "moderate"
}
It's worth noting that these slots are updated only because system provided that information in the previous message, if the system have not provided this information, the updated slots should be filled by "?"

"system": "Would you like me to book you a table?"
"user": "Not this time but I would like to visit an attraction while in Centre, can you find one for me?"
The "slots" updated by the "user" in this response are the following:

Copy code
{
    "restaurant-book_people": "dontcare",
    "restaurant-book_day": "dontcare",
    "restaurant-book_time": "dontcare",
    "attraction-area": "centre"
}
It's worth noting that restaurant-book_people,restaurant-book_day,restaurant-book_time are updated with "dontcare" since the user doesn't care about booking a table.

"system": "Okay, there's about 44 attractions in the centre of town.  Is there something in particular you'd like?"
"user": "no, but I will need the address."
The "slots" updated by the "user" in this response are the following:

Copy code
{
    "attraction-type": "?",
    "attraction-name": "?"
}
In this response the user didn't give any specific attraction details, so the "slots" of attraction-type and attraction-name are filled with "?" as the user didn't specify any preferences.

"system": "I think you would enjoy Old Schools in the centre of town.  It is an architectural attraction.  The address is trinity lane."
"user": "Great, that sounds perfect. Thanks!"
The "slots" updated by the "user" in this response are the following:

Copy code
{
    "attraction-type":"architectural",
    "attraction-name":"Old Schools",
    "attraction-address":"trinity lane"
}
The user is indicating that this attraction is the one they are interested in and thank the system, thus the slots are updated with the provided information.

"system": "Post code is cb21tt, Trinity lane. Is there anything else I can tell you?"
"user": "Not today , thank you so much for your help. I appreciate everything you do."
The "slots" updated by the user in this response are the following:

Copy code
{
    "attraction-postcode":"cb21tt"
}
The user is indicating the end of their conversation and expressing gratitude, so the system does not receive any new request.

"system": "Absolutely positively no problem at all! Thanks for using our service. Enjoy your time!"
"user": "Thanks, I'm all set here. You can end the conversation now."
There were no "slots" updated by the user in this response. The user is indicating the end of the conversation. Thus the response should be an empty JSON list {}
\end{lstlisting}

\section{Error Types}
\label{app:errors}

\subsection{Failed carry-over of system-informed values}
From \texttt{PMUL4050}:
\begin{lstlisting}
"system": "I'd recommend the Autumn House. Would you like to make a booking?"
"user": "Yes please. I need the reservation to be for 8 people and 2 nights starting on Tuesday. "
{
"hotel-book_people":"8",
"hotel-book_stay":"2",
"hotel-book_day":"Tuesday"
}

"system": "Great. I successfully booked your stay. Your reference number is 2OVWNFW1. Is there anything else I can help you with?"
"user": "I am looking for a particular attraction. Its name is called lynne strover gallery"
{
"attraction-name": "lynne strover gallery",
"attraction-type":"?"
}

"system": "The Lynne Strover Gallery is a wonderful museum in the West area of town, it is free to enter.  Would you like the phone number and address perhaps?"
"user": "Yes, I would also like to know what kind of attraction it is and the entrance fee."
{
"attraction-type": "?",
"attraction-fee":"?"
}

"system": "It is a museum and it is free entry."
"user": "What area is it in?"
{
"attraction-area":"west"
}

"system": "As stated previously, it is on the west side of town. Is there anything else I can do for you?"
"user": "No that's all the help I need. Thanks for your help. "
{}
\end{lstlisting}

\subsection{Incomplete coreference resolution}
From \texttt{MUL1228}:
\begin{lstlisting}
"system": "I have one hotel in the moderate range, cityroomz. Would you like me to book it for you?"
"user": "Yes, please. Can you book a room for Friday for 1 person, 3 nights?"
{
"hotel-name": "cityroomz",
"hotel-book_people": "1",
"hotel-book_day": "Friday",
"hotel-book_stay": "3"
}

"system": "Your reservation has been made. Reference number DTONYTIH. Is there anything else I can help you with?"
"user": "I need to book a taxi to get me to the restaurant by 21:30."
{
"taxi-destination": "?",
"taxi-arriveBy": "21:30",
"taxi-leaveAt": "dontcare",
"taxi-departure": "dontcare"
}

"system": "Will you be departing from your hotel?"
"user": "No, I'd like to leave the restaurant by 21:30, I need the taxi to take me to the hotel."
{
"taxi-destination": "hotel",
"taxi-leaveAt": "21:30",
"taxi-arriveBy": "dontcare",
"taxi-departure": "restaurant"
}
\end{lstlisting}

\subsection{Overprediction of \emph{dontcare}}
From \texttt{SNG01873}:

\begin{lstlisting}
"system": ""
"user": "I need to be picked up from pizza hut city centre after 04:30"

{
"taxi-leaveAt": "04:30",
"taxi-destination": "pizza hut city centre",
"taxi-departure": "?",
"taxi-arriveBy": "dontcare",
"restaurant-book_people": "dontcare",
"restaurant-book_day": "dontcare",
"restaurant-book_time": "dontcare",
"restaurant-food": "dontcare",
"restaurant-pricerange": "dontcare",
"restaurant-name": "dontcare",
"restaurant-area": "dontcare",
"hotel-book_people": "dontcare",
"hotel-book_day": "dontcare",
"hotel-book_stay": "dontcare",
"hotel-name": "dontcare",
"hotel-area": "dontcare",
"hotel-parking": "dontcare",
"hotel-pricerange": "dontcare",
"hotel-stars": "dontcare",
"hotel-internet": "dontcare",
"hotel-type": "dontcare",
"attraction-type": "dontcare",
"attraction-name": "dontcare",
"attraction-area": "dontcare",
"train-book_people": "dontcare",
"train-leaveAt": "dontcare",
"train-destination": "dontcare",
"train-day": "dontcare",
"train-arriveBy": "dontcare",
"train-departure": "dontcare"
}
\end{lstlisting}

\subsection{Ignoring value candidates}
From \texttt{MUL0540}:

\begin{lstlisting}
"system": "Booking was successful, the total fee is 188.8 GBP payable at the station. Reference number is: 31K4BK2E. Is there anything else you need?"
"user": "Yes, I am also looking for a multiple sports attraction in the city centre. Are you aware of any?"
{
"attraction-type":"multiple sports",
"attraction-area":"city centre"
}

"system": "Unfortunately, there are no multiple sports attractions in the area you specified. Is there some other type of attraction or area I can look into for you?"
"user": "Alright, that's fine, how about colleges? Are there any of those in the city centre?"
{
"attraction-type":"colleges",
"attraction-area":"city centre"
}
\end{lstlisting}

\subsection{Hallucinated slots}

From \texttt{MUL0671}:
\begin{lstlisting}
"system": "Cambridge has 8 moderately priced guesthouses in the north.  I would suggest Acorn Guest House, with a star rating of 4.  Would you like me to book it?"
"user": "Yes please, I would like to book it for 7 people for 5 nights on Saturday, and I need a reference number."
{
"hotel-name":"Acorn Guest House",
"hotel-stars":"4",
"hotel-book_people":"7",
"hotel-book_day":"Saturday",
"hotel-book_stay":"5",
"hotel-reference_number":"?"
}

"system": "I'm very sorry I was unable to book at acorn guest house for 5 nights, would you like to try for a shorter stay?"
"user": "Yes, please try to book it for four nights instead."
{
"hotel-book_stay":"4"
}

"system": "Your booking was successful, the reference number is WAEYAQ0M.  May I assist you with anything else today?"
"user": "No, that should be all for now. Thank you for the help!"
{
"hotel-reference_number":"WAEYAQ0M"
}
\end{lstlisting}

\subsection{Arbitrary normalization}

From \texttt{MUL0524}:
\begin{lstlisting}
"system": "The cambridge and county folk museum is located at 2-3 castle street postcode cb30aq , it has an entrance fee of 3.50 pounds. "
"user": "Thanks.  Now I need to book a train from Norwich."
{
"train-departure": "Norwich",
"train-destination": "?",
"train-leaveAt": "?",
"train-day": "?",
"train-arriveBy": "?",
"train-book_people":"?"
}

"system": "There are 133 trains leaving norwich. What day and time are you interested in?"
"user": "I'm going to Cambridge on saturday and want to arrive by 14:15 please."
{
"train-destination": "Cambridge",
"train-day": "Saturday",
"train-arriveBy": "14:15"
}
\end{lstlisting}

\subsection{Predicting $DS_t$ instead of $\widehat{DS}_t$}
From \texttt{MUL2116}:
\begin{lstlisting}
"system": ""
"user": "Yes, can you tell me information on Cambridge lodgings?  We are looking for free wifi and a nice guesthouse."
{
"hotel-name": "?",
"hotel-area": "?",
"hotel-parking": "dontcare",
"hotel-pricerange": "dontcare",
"hotel-stars": "dontcare",
"hotel-internet": "yes",
"hotel-type": "guest house"
}

"system": "Sure. What area and price range are you looking for?"
"user": "I don't really care about area or price range, the most important thing for me is free wifi in a guesthouse that has free parking."
{
"hotel-name": "?",
"hotel-area": "dontcare",
"hotel-parking": "yes",
"hotel-pricerange": "dontcare",
"hotel-stars": "dontcare",
"hotel-internet": "yes",
"hotel-type": "guest house"
}

"system": "I've got over 20 guesthouses that will work with those requirements.  Would you like me to book a moderately priced 4 star? "
"user": "As long as it has free wifi and parking then yes."
{
"hotel-name": "?",
"hotel-area": "dontcare",
"hotel-parking": "yes",
"hotel-pricerange": "moderate",
"hotel-stars": "4",
"hotel-internet": "yes",
"hotel-type": "guest house"
}

"system": "I would recommend the Acorn Guest House on the north side of town.  It has 4-stars.  Should I book this?"
"user": "Yes. Please book it for 8 people for 5 nights starting from sunday."
{
"hotel-name": "Acorn Guest House",
"hotel-area": "north",
"hotel-parking": "yes",
"hotel-pricerange": "moderate",
"hotel-stars": "4",
"hotel-internet": "yes",
"hotel-type": "guest house",
"hotel-book_people": "8",
"hotel-book_day": "sunday",
"hotel-book_stay":"5"
}
\end{lstlisting}

\section{``ChatGPT, are you a solution or an opportunity?''}
\label{app:question}

We queried ChatGPT with our titular question and received the following response at the 5th regeneration.

\begin{lstlisting}
ChatGPT for Zero-shot Dialogue State Tracking: A Solution or an Opportunity?

ChatGPT is a powerful language generation model that has the ability to generate human-like responses to a given prompt. It has been used in various natural language processing (NLP) tasks such as question answering, text summarization, and dialogue generation.

One potential application of ChatGPT is in zero-shot dialogue state tracking, which involves tracking the state of a conversation without any prior knowledge of the domain or the specific task. This could be useful in situations where the system needs to handle a wide range of topics and tasks, and where it is not feasible to manually annotate a large amount of training data.

However, there are also challenges with using ChatGPT for zero-shot dialogue state tracking. For example, it may not be able to accurately track the state of the conversation if the input is ambiguous or if the user's intent is not clear. Additionally, the model's responses may not always be appropriate or relevant to the current conversation.

Overall, while ChatGPT has the potential to be a solution for zero-shot dialogue state tracking, it also presents opportunities for further research and development in this area. It could be combined with other techniques such as reinforcement learning or active learning to improve its performance, or integrated with other NLP models such as dialogue state tracking models to provide more accurate and context-aware responses. 
\end{lstlisting}

\begin{figure*}
    \centering
    \lstset{basicstyle=\ttfamily\tiny, breaklines=true, postbreak=\mbox{\textcolor{red}{$\hookrightarrow$}\space}}
\begin{lstlisting}
Consider the following list of concepts, called "slots" provided to you as a json list.

"slots": {   
    "taxi-leaveAt": "the departure time of the taxi",
    "taxi-destination": "the destination of the taxi",
    "taxi-departure": "the departure of the taxi",
    "taxi-arriveBy": "the arrival time of the taxi",
    "restaurant-book_people": "the amount of people to book the restaurant for",
    "restaurant-book_day": "the day for which to book the restaurant",
    "restaurant-book_time": "the time for which to book the restaurant",
    "restaurant-food": "the food type of the restaurant",
    "restaurant-pricerange": "the price range of the restaurant",
    "restaurant-name": "the name of the restaurant",
    "restaurant-area": "the location of the restaurant",
    "hotel-book_people": "the amount of people to book the hotel for",
    "hotel-book_day": "the day for which to book the hotel",
    "hotel-book_stay": "the amount of nights to book the hotel for",
    "hotel-name": "the name of the hotel",
    "hotel-area": "the location of the hotel",
    "hotel-parking": "does the hotel have parking",
    "hotel-pricerange": "the price range of the hotel",
    "hotel-stars": "the star rating of the hotel",
    "hotel-internet": "does the hotel have internet",
    "hotel-type": "the type of the hotel",
    "attraction-type": "the type of the attraction",
    "attraction-name": "the name of the attraction",
    "attraction-area": "the area of the attraction",
    "train-book_people": "the amount of people to book the train for",
    "train-leaveAt": "the departure time of the train",
    "train-destination": "the destination of the train",
    "train-day": "the day for which to book the train",
    "train-arriveBy": "the arrival time of the train",
    "train-departure": "the departure of the train"
}

Some "slots" can only take a value from predefined list:

"categorical": {
    "hotel-pricerange": ["cheap", "moderate", "expensive"],
    "hotel-area": ["north", "south", "east", "west", "centre"],
    "hotel-parking": ["yes", "no"],
    "hotel-internet": ["yes", "no"],
    "hotel-type": ["hotel", "guest house"],
    "restaurant-pricerange": ["cheap", "moderate", "expensive"],
    "restaurant-area": ["north", "south", "east", "west", "centre"],
    "attraction-area": ["north", "south", "east", "west", "centre"]
}

Now consider the following dialogue between two parties called the "system" and "user". Can you tell me which of the "slots" were updated by the "user" in its latest response to the "system"? Present the updates in JSON format. If no "slots" were updated, return an empty JSON list. If you encounter "slots" that were requested by the "user" then fill them with "?". If a user does not seem to care about a discussed "slot" fill it with "dontcare".
\end{lstlisting}
    \caption{Prompt $P$.}
    \label{fig:unabridged_prompt_P}
\end{figure*}

\end{document}